\documentclass[11pt,a4paper]{rimlab}

\usepackage[utf8]{inputenc}
\usepackage[T1]{fontenc}
\usepackage{mathptmx}
\usepackage[margin=1in]{geometry}

\usepackage{graphicx}
\usepackage{amsmath,amssymb}
\usepackage{booktabs}
\usepackage{hyperref}
\usepackage{xcolor}
\usepackage[numbers,sort&compress]{natbib}

\newcommand{\clockMHz}{166}
\newcommand{\totalCyc}{35{,}107}

\title{Near-sensor Computing for Rapid Visuotactile Perception}

\author{Zhengying Zhu,
        Ruilin Zhang,
        Runze Hu,
        Chenxi Xiao\textsuperscript{*}\\[4pt]
        \small School of Information Science and Technology,
        ShanghaiTech University, \\Shanghai 201210, China\\
        \small\textsuperscript{*}Corresponding author: \texttt{xiaochx@shanghaitech.edu.cn}}
\date{}

\begin{document}

\begin{abstract}
Visuotactile sensors reconstruct dense contact geometry from measured surface gradients, but host-based processing increases power consumption and introduces data-transfer delays and variable scheduling latency, limiting the sensing and response speed of robotic systems. To address these limitations, we implement a near-sensor computing framework that includes a spectral Poisson solver as a fully streaming hardware pipeline. The computational core logic has an estimated power consumption of 347\,mW and achieves high throughput without data-dependent branching or iterative convergence, thereby providing deterministic latency. Operating at 166\,MHz, the pipeline produces the first depth value of each $128 \times 128$ frame 35{,}107 cycles after receiving the first input pixel, corresponding to a fixed latency of 0.211\,ms. Across 15 contact geometries, the reconstructed depths differ from a double-precision reference by 0.17\% of the peak contact depth. On-chip decisions based on these reconstructions close a robot protective reflex loop in 28.3\,$\pm$\,4.9\,ms, compared with 169.9\,$\pm$\,27.8\,ms for an equivalent host-based loop using the same actuator. These results demonstrate that near-sensor reconstruction can provide accurate, energy-efficient, and deterministic tactile geometry on timescales suitable for rapid robotic contact responses.
\end{abstract}

\maketitle

Visuotactile sensors provide particularly rich spatial information compared with other tactile sensing modalities~\cite{yuan2017gelsight,lambeta2020digit,taylor2022gelslim3,li2024signalprocessing,luo2025tactilerobotics,qin2026nlipscalib, lin2025pp, wu2025humanft}. By imaging the deformation of a soft elastomer under structured illumination, these sensors reconstruct dense three-dimensional (3D) contact geometry using photometric stereo and Poisson integration~\cite{yuan2017gelsight,li2014gelsight,lambeta2020digit,wu2025humanft,qin2026nlipscalib}. The resulting depth maps enable the estimation of a wide range of physically meaningful quantities, including contact-force distributions~\cite{ma2019densetactile}, surface curvature~\cite{lin2023_9dtact}, incipient slip~\cite{dong2017improved}, and object pose~\cite{suresh2024neuralfeels,zhao2026gelslam}. Together, these capabilities make visuotactile sensing particularly well suited for superhuman tactile perception and dexterous robotic manipulation.

Despite their rich spatial representation, current visuotactile sensors are still too slow for many high-speed robotic tasks. The principal bottleneck is 3D reconstruction: recovering contact geometry typically requires solving a Poisson partial differential equation over dense gradient fields, a computation that is substantially more expensive than the processing used in most other tactile modalities. On general-purpose CPUs, reconstruction often takes tens of milliseconds~\cite{yuan2017gelsight}, with further variability arising from operating-system scheduling and host-based data transfer. This latency reduces control bandwidth and temporal determinism when geometric feedback is required~\cite{lee2025highbandwidth}, limiting the use of visuotactile sensing in rapid slip reaction, dynamic manipulation, and other ultrafast closed-loop behaviours.

Existing efforts to mitigate this bottleneck have largely followed two directions, each involving a fundamental trade-off. One line of work improves throughput by simplifying the representation, for example by estimating surface normals or directly predicting task-specific variables from raw tactile images instead of reconstructing full 3D geometry~\cite{li2024signalprocessing,kota2026_3dcal}. Such approaches can achieve update rates of around 100-200~Hz in recent systems, but they still do not provide a unified 3D geometric description of contact that can be reused across tasks. A second direction pursues hardware-level acceleration through event-based tactile sensing, which replaces conventional frame-based imaging with event-based tactile sensors and can reach kilohertz-level temporal resolution~\cite{funk2024evetac,jiang2026spikingtac}. However, because these signals are sparse and encode only temporal changes rather than dense surface structure, such systems are mainly suited to coarse perceptual tasks such as contact detection or classification, and remain unsuitable for dense geometric reconstruction such as depth estimation~\cite{funk2024evetac}.

More recently, near-sensor and in-sensor computing paradigms have been explored in tactile sensing as a way to reduce delay and improve efficiency~\cite{zhou2020nearsensor}. Prior studies have demonstrated edge-side processing for low-level operations such as filtering or feature extraction in capacitive~\cite{chen2025capacitive,cho2025npjflex}, piezoresistive \cite{leng2025touch}, and other tactile sensing platforms, highlighting the value of moving simple signal processing closer to the sensor.
However, extending this strategy to visuotactile sensing is substantially more challenging. The central difficulty is that useful output requires dense 3D surface reconstruction, which in turn demands solving a Poisson equation at the sensor interface in real time.
In parallel, hardware-accelerated PDE solvers have been developed for application domains unrelated to tactile perception, such as scientific computing and medical imaging~\cite{kamalakkannan2021fpgastencil,tomczak2023nsfpga,elaraby2007multigridfpga,siddiqui2017fpgasense,omer2020fpgasense}. Many of these architectures use iterative numerical methods, including multigrid, Jacobi and conjugate-gradient solvers, whose iteration counts depend on the input field and the convergence tolerance; their latency is therefore not fixed by construction.
By contrast, robotic visuotactile perception demands dense reconstruction at the camera frame rate, with deterministic low latency and within a tightly constrained power budget. Moving the Poisson solve onto the sensor node is therefore an active direction to explore.

To address this gap, we present a near-sensor visuotactile computing architecture that integrates sensing, dense reconstruction, and contact-triggered decision-making at the sensor node (Fig.~\ref{fig:system}). Following the framework in \cite{xiao_patent}, the architecture couples a visuotactile sensor operating at up to 400~frames~s$^{-1}$ with an on-chip spectral Poisson solver at $128\times128$ resolution. Rather than accelerating an iterative procedure, the solver maps the direct spectral formulation onto a fixed-schedule streaming datapath, ensuring an identical execution path for every frame. At 166\,MHz, the first reconstructed depth value is produced with a fixed latency of 0.211\,ms, while the computational core consumes an estimated 324\,mW. The hardware reconstruction deviates from a double-precision reference by only 0.17\% of the peak contact depth and enables a protective reflex approximately six times faster than an equivalent host-based implementation.

\section*{Results}
\subsection*{Near-sensor architecture for ultrafast visuotactile sensing}

\begin{figure}[!ht]
\centering
\includegraphics[width=\textwidth]{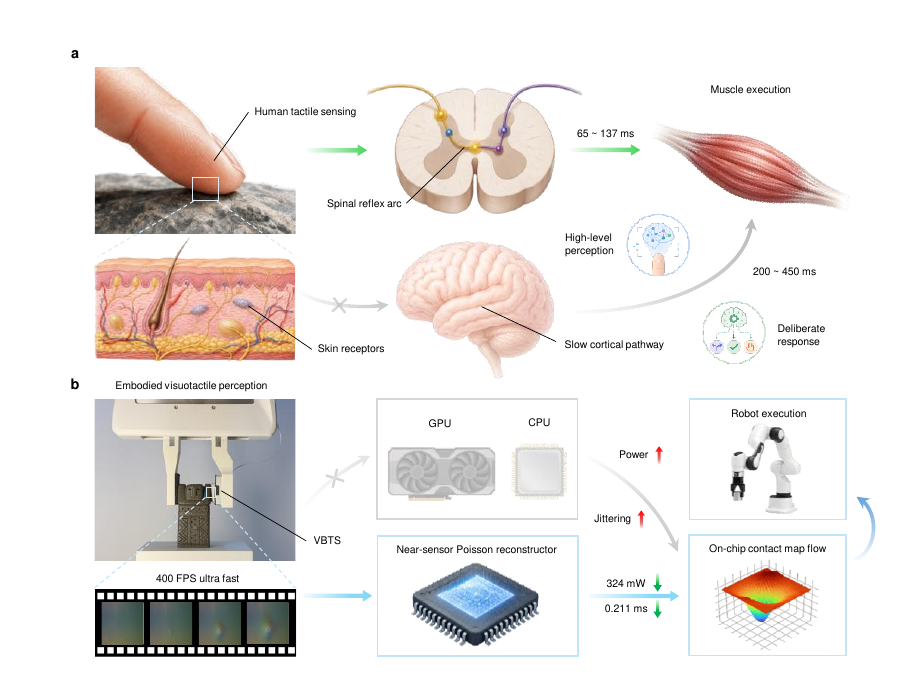}
\caption{\textbf{Deterministic near-sensor visuotactile perception.}
\textbf{a,} Biological protective reflex. A cutaneous stimulus is relayed to the spinal cord, where a protective response is initiated within 65--137\,ms~\cite{thorell2024withdrawal}; the slower cortical pathway supports deliberate responses on a 200--450\,ms timescale.
\textbf{b,} Engineered counterpart. Frames from the ultrafast vision-based tactile sensor (VBTS) are converted into dense depth maps by the near-sensor processor and acted upon locally. The first depth value is produced at a fixed latency of 0.211\,ms after the first input pixel arrives, followed by one depth value per clock cycle, consuming 347\,mW (FPGA programmable logic estimate), and 324\,mW (45\,nm ASIC estimate). This local pathway avoids the additional power consumption, data-transfer delay, and timing variability of host-based CPU or GPU processing.}
\label{fig:system}
\end{figure}

High-resolution visuotactile sensors typically rely on imaging arrays to capture the deformation of a soft elastomer. Under structured illumination, a prominent class of these sensors essentially measures the local surface gradient at each point rather than depth directly~\cite{yuan2017gelsight, lambeta2020digit, qin2026nlipscalib, wu2025humanft}.
Recovering the actual 3D shape of the contact requires mathematically integrating these gradient vectors across the entire sensor.
Because optical measurements are inherently noisy, direct path-wise integration accumulates measurement noise and yields path-dependent artefacts. Finding the optimal smooth surface becomes a global integration problem, formulated as solving a Poisson partial differential equation ($\nabla^2 z = \nabla \cdot \mathbf{g}$).
In conventional setups, this means streaming raw RGB video to a host processor, where the depth map is reconstructed in software by a direct transform-based Poisson solve~\cite{yuan2017gelsight,wang2021gelsightwedge}. However, sending the raw video to a host incurs transfer latency and additional power consumption~\cite{chen2025capacitive}, and operating-system scheduling makes the reconstruction time vary from frame to frame.

Fig.~\ref{fig:system} presents the proposed visuotactile perception system. Its local sensor--processor--actuator pathway resembles a biological protective reflex, in which withdrawal is initiated without waiting for cortical processing (Fig.~\ref{fig:system}a). The system combines a visuotactile sensor operating at up to 400~frames~s$^{-1}$ with a low-latency, power-efficient architecture that reconstructs dense depth maps directly from the incoming pixel stream, without host intervention (Fig.~\ref{fig:system}b). Unlike conventional CPU- or GPU-based processing, which requires off-sensor data transfer and is subject to variable scheduling delays, the near-sensor architecture performs reconstruction on-chip with deterministic latency.

The architecture is built around a spectral Poisson solver, which exploits the eigen-decomposition of the discrete Laplacian under Dirichlet boundary conditions.
Specifically, the two-dimensional Poisson equation $\nabla^2 z = \nabla \cdot \mathbf{g}$ reduces to element-wise division in the spectral domain after a discrete sine transform (DST), yielding a direct solution of the discrete system in a fixed number of exact arithmetic operations~\cite{strang2007cse} (see Methods and Supplementary Section~S1).
Software reconstruction pipelines already solve the equation this way, by a direct transform rather than by iteration~\cite{yuan2017gelsight}; what has not been exploited is a fully pipelined circuit in which latency is fixed by construction. In contrast, hardware PDE accelerators have generally been built around iterative solvers, whose convergence time depends on the input field and the target accuracy~\cite{elaraby2007multigridfpga,kamalakkannan2021fpgastencil}.

The complete pipeline is implemented as a streaming datapath on a Xilinx Zynq UltraScale+ FPGA. Pixels enter in raster-scan order at the camera pixel rate, with no full-frame input buffering. A photometric stereo core converts the raw RGB stream into per-pixel surface gradients under the tri-colour model. A Poisson core integrates those gradients into a dense depth map by the spectral method, with each one-dimensional DST computed by a pipelined FFT core with input reordering and phase correction~\cite{makhoul1980fct} and the two-dimensional transform performed separately.
Because the cores form a single feed-forward pipeline, gradient estimation and Poisson integration proceed concurrently on successive regions of each frame, so the end-to-end latency is fixed and independent of contact geometry.
In the following sections, we characterize three aspects of this architecture in turn: its deterministic latency and power efficiency, the fidelity of the reconstructed geometry, and its use as a low-latency tactile front end.

\subsection*{Deterministic latency and power efficiency}

\begin{figure}[!ht]
\centering
\includegraphics[width=\textwidth]{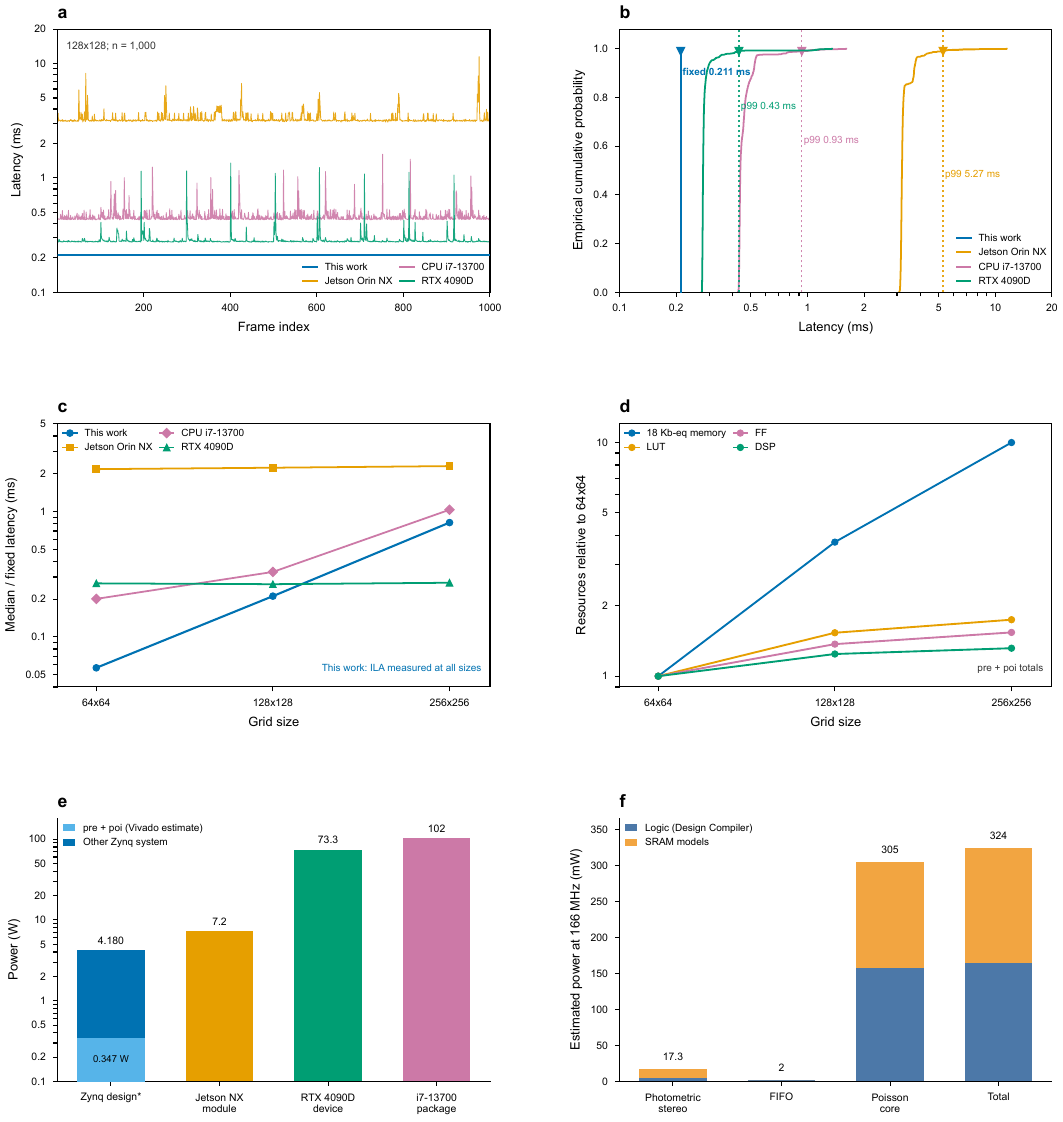}
\caption{\textbf{Deterministic latency, scaling, and power.}
\textbf{a,} Per-frame reconstruction latency over 1{,}000 consecutive $128\times128$ frames ($n=1{,}000$ for each platform).
\textbf{b,} Cumulative distribution of the latency measurements in \textbf{a}.
\textbf{c,} Reconstruction latency as a function of grid size. Near-sensor results are on-chip measurements, whereas software results show the median latency at each size.
\textbf{d,} On-chip resource usage as a function of grid size, summed across both cores and normalized to the $64\times64$ implementation. Memory usage is reported in equivalent 18\,Kb blocks.
\textbf{e,} Power consumption of each platform, with the corresponding measurement boundary indicated on each bar. 
The asterisk marks a Vivado simulation estimate; the other three are measured. The two reconstruction cores consume 347~mW according to the Vivado simulation estimate.
\textbf{f,} Estimated ASIC power consumption at 166\,MHz, separated into synthesized logic and modelled SRAM contributions.}
\label{fig:latency}
\end{figure}

Robotic feedback control requires perception latency that is not only low but predictable.
Therefore, we studied reconstruction latency, which is the interval from the first pixel of a frame entering the photometric stereo core to the first depth value leaving the Poisson core. Over 1{,}000 consecutive $128 \times 128$ frames this latency was 35{,}107~clock cycles at 166\,MHz for every frame, a fixed 0.211~ms (Fig.~\ref{fig:latency}a), so its latency distribution is a single line rather than a spread (Fig.~\ref{fig:latency}b).
Software implementations of the same solver do not share this property. To show this, the software reconstruction was run on an Intel Core~i7-13700 CPU, an NVIDIA RTX~4090~D GPU and an NVIDIA Jetson Orin~NX.
Running on shared-resource processors, these implementations produce latencies that vary from frame to frame, with long-tailed distributions: the 99th-percentile latency reaches 0.93\,ms on the desktop CPU, 0.43\,ms on the desktop GPU and 5.3\,ms on the embedded Jetson (Fig.~\ref{fig:latency}b). Latency at these grid sizes is dominated by software overheads rather than by the transform, which is why the desktop-GPU trace barely moves with grid size (Fig.~\ref{fig:latency}c).

In our system, throughput is determined by the pixel clock rather than by the latency of an individual frame. Because the pipeline accepts one pixel per clock cycle, a $128\times128$ frame requires 16{,}384 input cycles, giving a theoretical throughput limit of 10{,}133~frames~s$^{-1}$ at 166\,MHz. When synthetic frames were streamed continuously to the deployed system through its DMA interface, we measured 10{,}106~frames~s$^{-1}$ of correct depth output, within 0.3\% of the theoretical limit. The small difference is attributable to interrupt and logging overhead rather than to the solver itself. As the camera operates well below this rate, the deployed system is limited by sensor throughput rather than reconstruction throughput.

Latency is deterministic for a given grid size. Because the pipeline contains no iterative loops, the cycle count is fixed, resulting in $O(N^2)$ scaling for an $N\times N$ grid. Across 1{,}000 consecutive frames at each resolution, the measured latency was 9{,}421 cycles at $64\times64$, 35{,}107 cycles at $128\times128$, and 135{,}448 cycles at $256\times256$, corresponding to 0.057, 0.211, and 0.816\,ms at 166\,MHz, respectively. Every frame at a given resolution had the same cycle count (Fig.~\ref{fig:latency}c), while doubling the grid dimension increased the latency by approximately fourfold, consistent with quadratic scaling in number of pixels.

We also analysed how resource usage scales with grid size. Because the transform core is reused for the row and column passes, arithmetic resources grow only modestly: from $64\times64$ to $256\times256$, despite a sixteenfold increase in pixel count, DSP usage increases from 82 to 108 blocks and logic usage from 12{,}280 to 21{,}413 LUTs (Fig.~\ref{fig:latency}d). Storage requirements, by contrast, scale with frame size. The three implementations use 8, 32, and 64 of the device's 96 UltraRAM blocks, respectively. The $256\times256$ implementation additionally uses 176 block-RAM tiles because its frame buffers exceed the available UltraRAM capacity. Thus, the maximum supported grid size is constrained primarily by on-chip memory rather than arithmetic resources, and this limit depends on the capacity of the target device rather than on the architecture itself.

The near-sensor implementation also operates within a substantially lower power envelope. For the FPGA implementation, Vivado's simulation-based power analysis estimates approximately 4.180\,W for the complete Zynq design with the reconstruction cores operating at 166\,MHz. Most of this power is attributed to the processing system and peripheral blocks that are not involved in reconstruction; the estimate therefore characterizes the deployed prototype rather than the reconstruction datapath alone (Fig.~\ref{fig:latency}e). By comparison, sustaining reconstruction at 400\,frames\,s$^{-1}$ requires 102\,W of processor-package power on the desktop CPU and 73\,W of device power on the desktop GPU. The embedded Jetson consumes 7.2\,W of total module power at a maximum of approximately 180\,frames\,s$^{-1}$. Because the proposed architecture is not inherently tied to the FPGA substrate, we also evaluated its potential for future fingertip-scale integration. When synthesized in a 45\,nm ASIC process using Synopsys Design Compiler, the reconstruction datapath occupies 5.10\,mm$^2$ and has an estimated power consumption of 324\,mW, including on-chip SRAM, at the same 166\,MHz clock frequency (Fig.~\ref{fig:latency}f; Methods).

\subsection*{Reconstruction quality}

\begin{figure}[!ht]
\centering
\hspace{-30pt}
\includegraphics[width=\textwidth]{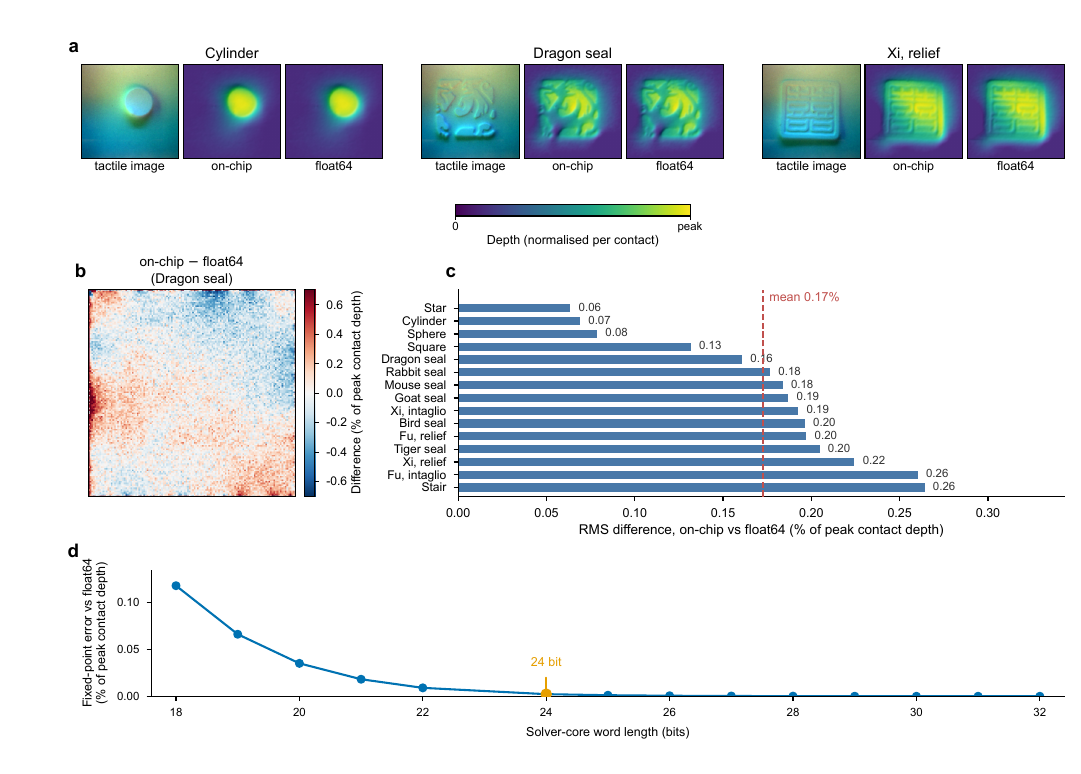}
\caption{\textbf{Reconstruction accuracy and fixed-point fidelity.}
\textbf{a,} Depth maps reconstructed on-chip, versus using the double-precision software reference for representative contact geometries, together with the corresponding raw tactile images. The shared colour scale is normalized separately by the peak depth of each contact.
\textbf{b,} Pixel-wise differences between the hardware and software reconstructions, with the residuals occurring mostly near the sensor boundaries and contact region. The colour scale shows the signed difference as a percentage of each contact's peak depth.
\textbf{c,} Agreement between hardware and software across all 15 tested geometries, quantified by the root-mean-square difference relative to peak contact depth after per-frame scale-and-offset alignment.
\textbf{d,} Fixed-point error of the solver core relative to the double-precision reference as a function of internal word length; the deployed 24-bit configuration is highlighted.}
\label{fig:accuracy}
\end{figure}

A further question is whether resource-constrained near-sensor computation compromises reconstruction accuracy. We therefore compared the on-chip results with a double-precision, host-based software reference computed from the same raw frames. The evaluation covered 15 contact geometries, including machined indenters and 3D-printed relief and intaglio seals (Fig.~\ref{fig:accuracy}a,b).

The 24-bit fixed-point datapath closely reproduces the software results. To account for calibration-gain differences, we aligned the hardware and software depth maps for each geometry using a frame-wise scale and offset. Across all 15 geometries, the root-mean-square difference averaged 0.17\% of the peak contact depth, with individual values ranging from 0.06\% to 0.26\% (Fig.~\ref{fig:accuracy}c), and the reconstructed depth maps were visually indistinguishable (Fig.~\ref{fig:accuracy}a). The pixel-wise residuals occurred mostly near the boundaries as well as the contact region (Fig.~\ref{fig:accuracy}b).

This balance between resource efficiency and reconstruction accuracy was achieved by optimizing the arithmetic precision at each stage. A word-length sweep of the solver core against a double-precision reference identified 24 bits as the optimal trade-off between accuracy and hardware resources. Accuracy degraded progressively with shorter word lengths but plateaued beyond 24 bits (Fig.~\ref{fig:accuracy}d). Accordingly, the deployed implementation uses 24-bit arithmetic throughout.

\subsection*{A physical reflex loop closed at the sensor}

\begin{figure}[!ht]
\centering
\includegraphics[width=\textwidth]{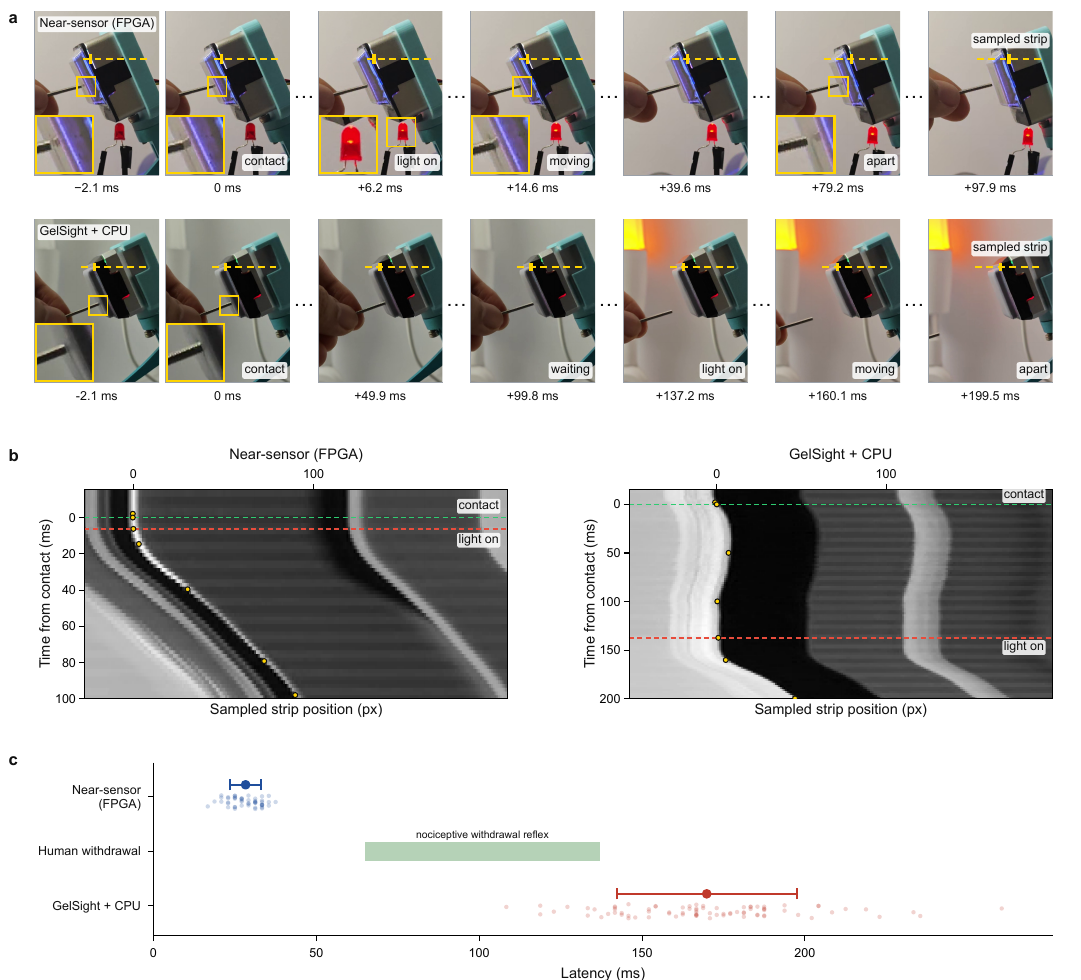}
\caption{\textbf{A protective reflex closed at the sensor.}
\textbf{a,} Representative high-speed camera sequences for the near-sensor FPGA and GelSight--host pathways, aligned to first contact. The dashed line indicates the sampled strip, and the tick marks the tracked edge of the housing.
\textbf{b,} Space--time views of the sampled strips, with time increasing downwards. The green and red dashed lines indicate contact and command onset, respectively, with command onset signalled by illumination of an LED.
\textbf{c,} Contact-to-motion latency for all near-sensor ($n=50$) and host-based ($n=81$) trials. Reference latency ranges for human nociceptive withdrawal reflexes are included for comparison~\cite{thorell2024withdrawal}. Dots represent individual trials; markers and error bars indicate mean $\pm$ s.d.}
\label{fig:reflex}
\end{figure}

Finally, we investigated whether the fast and deterministic sensor response could enable a robotic analogue of the human protective withdrawal reflex. We probed the sensor surface with a pin, using contact depth to represent increasing deformation of skin. In this setting, the robot needs to retract the finger within a safety window, analogous to a human withdrawing a finger before tissue damage occurs.

To implement this reflex, we integrated a threshold comparator into the programmable logic. The comparator evaluates each depth value as it exits the reconstruction pipeline. When the contact depth exceeds a programmable threshold, a GPIO signal is asserted within one clock cycle, directly triggering the motor command without a host processor in the control path. As a baseline, we reproduced a conventional host-based pipeline using the same actuator: a GelSight Mini visuotactile sensor streams data to a PC over USB, where depth is reconstructed and the threshold condition is evaluated in software before the motor command is sent. Both pathways drive the same servo assembly and therefore share the mechanical portion of the loop, allowing the comparison to isolate the latency associated with sensing, reconstruction, communication, and decision-making (Fig.~\ref{fig:reflex}a,b).

An external high-speed camera operating at 480\,frames\,s$^{-1}$ resolved the onset of contact, command and motion in each trial. The contact-to-command latency was 10.2\,$\pm$\,3.2\,ms for the near-sensor pathway, compared with 148.6\,$\pm$\,28.3\,ms for the host-based loop, corresponding to an approximately 14.6-fold reduction. The complete contact-to-motion latency was 28.3\,$\pm$\,4.9\,ms for the near-sensor pathway ($n=50$; median, 29.2\,ms; range, 16.7--37.5\,ms), compared with 169.9\,$\pm$\,27.8\,ms for the host-based loop ($n=81$; median, 166.7\,ms; range, 108.3--260.4\,ms). This represents a 6.0-fold reduction in latency (two-sided Mann--Whitney $U=0$, $P=8.1\times10^{-22}$; values are mean~$\pm$~s.d.) (Fig.~\ref{fig:reflex}c).

Notably, 18.1\,$\pm$\,3.3\,ms of the total 28.3\,ms contact-to-motion latency is attributable to the servo response rather than to sensing and decision-making. The closest biological analogue is the human nociceptive withdrawal reflex, for which latencies of 65--137\,ms have been reported~\cite{thorell2024withdrawal}. The near-sensor loop (28.3\,ms) falls below the reported range, whereas the loop through the host computer (169.9\,ms) lies well above it.

\section*{Discussion}

The main technical contribution of this work is mapping a Poisson solver onto a fully pipelined hardware architecture. By solving the Poisson equation on-chip, the sensor outputs calibrated depth in millimetres rather than raw pixels. Depth provides a common representation for downstream tasks, enabling force-estimation models to transfer across visuotactile sensors~\cite{chen2026genforce} and improving tactile SLAM over appearance-based features~\cite{zhao2026gelslam}. Unlike iterative solvers, our pipeline follows a fixed sequence of operations without data-dependent control flow, providing deterministic latency and making PDE-based near-sensor processing practical.

Previous near- and in-sensor tactile systems have focused mainly on filtering~\cite{chen2025capacitive}, edge enhancement~\cite{cho2025npjflex} and classification~\cite{zhou2020nearsensor}. A concurrent preprint~\cite{lu2026fastac} reports a DST-based Poisson depth solver for a visuotactile sensor, implemented on an XCZU19EG FPGA and operating at the camera rate of 240\,Hz. Together, both that study and ours demonstrate the value of near-sensor visuotactile computation. Our implementation achieves lower reconstruction latency (0.211\,ms versus 0.87\,ms) and supports higher throughput and resolution on a much smaller FPGA, whereas the concurrent work emphasizes near-infrared illumination for improved reconstruction quality and three-axis force estimation. Our architecture is also complementary to event-based tactile sensors~\cite{funk2024evetac,jiang2026spikingtac}: event-based sensors capture rapid changes such as slip, vibration and texture edges, whereas our pipeline reconstructs dense 3D depth at high temporal bandwidth. We further demonstrate a protective reflex closed entirely at the sensor, with a contact-to-motion latency below the reported range for human nociceptive withdrawal reflexes.

An alternative would be to implement an iterative solver, such as multigrid or conjugate gradients, on the same FPGA. For a $128 \times 128$ grid, a multigrid V-cycle can converge in a few iterations in software, with each iteration requiring fewer operations than a full spectral transform. On an FPGA, however, iterative methods require repeated read--modify--write passes over the solution array. Convergence-based stopping therefore introduces variable latency, whereas fixing the iteration count requires either repeated passes through shared hardware or replication of the stencil hardware. Neither approach provides the same combination of streaming execution, deterministic latency and efficient resource use as this spectral pipeline. To our knowledge, this pipeline achieves the lowest reported latency for a Poisson solver at a comparable problem size~\cite{elaraby2007multigridfpga,kamalakkannan2021fpgastencil,tomczak2023nsfpga}.

A current limitation is the supported grid size. Although the streaming architecture is not limited to the deployed $128 \times 128$ configuration, and both $64 \times 64$ and $256 \times 256$ implementations were successfully demonstrated on the same FPGA, grids larger than $256 \times 256$ are currently limited by the available on-chip memory for transform buffering. Future designs could overcome this limitation by incorporating external memory. More broadly, the streaming architecture is not specific to tactile reconstruction and could be extended to other scientific and engineering applications governed by Poisson equations.

\section*{Methods}
\subsection*{Sensor and hardware platform}

The details of our hardware platform are described as follows. Since high-speed visuotactile sensors are scarcely available commercially, we developed a visuotactile sensor based on GelSight~\cite{li2014gelsight}. The sensor consists of a soft elastomer membrane coated with a reflective layer, illuminated by red, green, and blue light sources, and imaged by a CMOS camera. The CMOS image sensor is designed to be interchangeable, allowing different trade-offs among spatial resolution, frame rate, region of interest (ROI), and field of view (FoV).

The sensor head is connected to a Xilinx Zynq UltraScale+ XCZU7EV system-on-chip via a MIPI CSI-2 interface, forming a self-contained sensing unit. All depth reconstruction is performed entirely within the FPGA programmable logic, operating at \clockMHz\,MHz. The on-chip ARM processing system runs an LwIP Ethernet stack solely for transmitting reconstructed depth frames to a host PC. It is not involved in the reconstruction pipeline or the reflex decision-making process.

\subsection*{Reconstruction pipeline}

Depth reconstruction solves the Poisson equation $\nabla^2 z = \nabla \cdot \mathbf{g}$ using a spectral method that diagonalizes the discrete Laplacian under homogeneous Dirichlet boundary conditions~\cite{strang2007cse}; the derivation is provided in Supplementary Section~S1. The computation decomposes into a fixed sequence of streaming operations: gradient lookup, finite differencing, forward transform, point-wise spectral division, and inverse transform, with no data-dependent branching or iterative convergence. Consequently, every frame traverses the pipeline in exactly \totalCyc{} clock cycles.

The datapath adopts a single-pass streaming architecture. Pixels enter in raster-scan order at the camera pixel rate and are processed on the fly. The photometric stereo stage maps the tri-colour intensities to per-pixel surface gradients through a lookup module. Each one-dimensional transform is implemented with a pipelined DST core followed by input reordering and phase correction~\cite{makhoul1980fct}, and the two-dimensional transform is computed separably, with the two passes connected by full-frame matrix transpositions. All arithmetic uses a 24-bit signed fixed-point representation. The precision sweep reported in Fig.~\ref{fig:accuracy}d varies the internal word length of the Poisson solver core in bit-accurate simulation and compares it against the same core configured with the double-precision reference.

\subsection*{ASIC synthesis}

The same RTL was synthesized in a 45\,nm standard-cell process (NanGate FreePDK45, typical corner, 1.1\,V, 25\,$^\circ$C) using Synopsys Design Compiler. The transposition buffers and the calibration ROM were modelled as single-port SRAM using CACTI and \texttt{bsg\_fakeram}. Because the FPGA implementation uses dual-port memory, the reported SRAM area is a lower bound. Switching activity was extracted from RTL simulation of recorded tactile frames over a steady-state window, annotated at the RTL ports and propagated internally by the synthesis tool without gate-level simulation. The resulting post-synthesis implementation occupies 5.10\,mm$^2$ and consumes 324\,mW, including SRAM, at the 166\,MHz operating clock.

\subsection*{Latency characterisation and software baselines}

FPGA per-frame latency was measured on-chip using the Vivado Integrated Logic analyser over 1{,}000 consecutive frames for each of the three grid sizes, by counting clock cycles from the arrival of the first pixel of a frame to the output of the first reconstructed depth value. Software baselines used a PyTorch implementation of the spectral Poisson reconstruction on individual $128 \times 128$ frames (batch size 1). The forward and inverse discrete sine transforms were constructed using \texttt{torch.fft.fft} and \texttt{torch.fft.ifft}. The workload was evaluated on an Intel Core~i7-13700 CPU, an NVIDIA GeForce RTX~4090~D GPU, and an NVIDIA Jetson Orin~NX~\cite{mittal2019jetsonsurvey}. For each platform, 100 warm-up iterations were discarded, followed by 1{,}000 timed per-frame measurements.

For the CPU and RTX~4090~D, the reported latency includes only the Poisson reconstruction computation, with the input data and the invariant Poisson denominator precomputed before timing. CPU latency was measured using \texttt{time.perf\_counter}. RTX~4090~D latency was measured using the same wall-clock timer, with CUDA synchronization immediately before and after each iteration. For the Jetson Orin~NX, latency was measured using \texttt{time.perf\_counter\_ns}. Since the Jetson is an integrated system-on-chip, the reported power consumption corresponds to the combined CPU--GPU package while CPU utilization was minimized.

\subsection*{Reconstruction accuracy}

Accuracy was evaluated on 15 contact geometries comprising indenters, relief seals, and intaglio seals. For each contact, the raw tactile frame and the hardware-generated depth map were captured simultaneously, and the same raw frame was reconstructed offline in double-precision software on the host computer using the identical calibration constants. Before comparison, a scale factor and an offset were fitted for each frame and applied to the software reconstruction to compensate for numerical discrepancies between the fixed-point hardware implementation and the double-precision software implementation. The hardware and software depth maps were then compared pixel-wise over the full frame excluding a one-pixel border. The root-mean-square difference normalized by the peak contact depth of each geometry is reported.

\subsection*{Reflex loop measurement}

The reflex loop was measured on a robotic finger driven by four serial-bus servomotors, all triggered by a single synchronised goal-position command. Two pathways were compared using the same actuator. In the near-sensor pathway, a threshold comparator implemented in the programmable logic evaluates each reconstructed depth value against a programmed register value, directly triggering the motor command. In the baseline pathway, a commercial visuotactile sensor (GelSight Mini) streams data over USB to a PC, where depth reconstruction is performed in software running on the host CPU. The same threshold operation is applied, and the motor command is issued by the host.

Timing was recorded using an external high-speed camera at 480\,frames\,s$^{-1}$ (frame interval 2.083\,ms), which simultaneously observed contact, command issuance, and the onset of finger motion within the same field of view. An LED indicator driven by the command bus marks the instant at which the command is issued. Each trial was analysed frame by frame to obtain $t_1$, the interval from contact to command issuance, and $t_2$, the interval from command issuance to visible motion onset; the complete reflex loop latency is therefore $t_1+t_2$. The measured latency values are quantized by the 2.083\,ms camera frame interval, which defines the temporal resolution of the comparison. Trials were pooled across multiple recording sessions, comprising 50 trials for the near-sensor pathway and 81 trials for the host-loop baseline. Group differences were assessed using a two-sided Mann--Whitney $U$ test.

\section*{Acknowledgements}

This work was supported by the Natural Science Foundation of Shanghai (Grant No. 25ZR1402370) and received partial support from the Key Laboratory of Intelligent Perception and Human-Machine Collaboration (ShanghaiTech University), Ministry of Education.

\section*{Author contributions}

Z.Z. developed the methodology and performed experiments on FPGA. Z.Z., R.Z., and C.X. analysed the data, drafted the manuscript, and interpreted the results. R.Z. contributed to the sensor design and conducted experiments on baseline devices. R.H. developed the mechanism for the reflex experiment. C.X. initiated and conceived the project, supervised the research.
All authors reviewed and approved the final version.

\bibliographystyle{naturemag}
\bibliography{reference}

\clearpage

\section*{Supplementary Information}

\setcounter{equation}{0}
\renewcommand{\theequation}{S\arabic{equation}}
\setcounter{figure}{0}
\renewcommand{\thefigure}{S\arabic{figure}}
\setcounter{table}{0}
\renewcommand{\thetable}{S\arabic{table}}

\subsection*{S1.\quad Spectral diagonalisation of the discrete Poisson equation}

This section derives the spectral solver used in the depth reconstruction pipeline. The treatment follows Strang~\cite{strang2007cse}, Sections~1.5 and~3.5; we reintroduce the key steps here for completeness.

\paragraph{Problem statement.}
The depth map $z$ is obtained by solving the two-dimensional Poisson equation on an $N \times N$ grid under homogeneous Dirichlet boundary conditions:
\begin{equation}
  \nabla^2_d \, z_{i,j} = f_{i,j}, \qquad i,j = 1,\ldots,N, \qquad z = 0 \;\text{on the boundary},
  \label{eq:s_poisson}
\end{equation}
where $f_{i,j} = (\nabla \cdot \mathbf{g})_{i,j}$ is the divergence of the surface-gradient field computed from the photometric lookup table, and $\nabla^2_d$ denotes the standard five-point discrete Laplacian:
\begin{equation}
  \nabla^2_d \, z_{i,j} = z_{i-1,j} + z_{i+1,j} + z_{i,j-1} + z_{i,j+1} - 4\,z_{i,j}.
  \label{eq:s_laplacian}
\end{equation}

\paragraph{1D eigenvalue problem.}
We first consider the one-dimensional case. The $N \times N$ tridiagonal matrix
\begin{equation}
  T_N =
  \begin{pmatrix}
    -2 &  1 &        &    \\
     1 & -2 & \ddots &    \\
       & \ddots & \ddots & 1 \\
       &        &  1     & -2
  \end{pmatrix}
  \label{eq:s_tridiag}
\end{equation}
arises from the 1D Laplacian with Dirichlet boundary conditions ($z_0 = z_{N+1} = 0$). Its eigenvalues and eigenvectors are well known~\cite{strang2007cse}:
\begin{align}
  \lambda_k &= -2 + 2\cos\frac{k\pi}{N+1} = 2\left(\cos\frac{k\pi}{N+1} - 1\right), \qquad k = 1,\ldots,N,
  \label{eq:s_eigenval}\\[4pt]
  [\mathbf{s}_k]_j &= \sin\frac{jk\pi}{N+1}, \qquad j = 1,\ldots,N.
  \label{eq:s_eigenvec}
\end{align}
The eigenvectors $\{\mathbf{s}_k\}$ are orthogonal and form the basis of the discrete sine transform (DST). Concretely, the matrix $S$ with entries $S_{jk} = \sin(jk\pi/(N{+}1))$ diagonalizes $T_N$:
\begin{equation}
  T_N = S \,\Lambda\, S^{-1}, \qquad \Lambda = \mathrm{diag}(\lambda_1,\ldots,\lambda_N), \qquad S^{-1} = \tfrac{2}{N+1}\,S^\top.
  \label{eq:s_diag}
\end{equation}

\paragraph{Extension to 2D.}
The 2D discrete Laplacian in Eq.~\eqref{eq:s_laplacian} can be written as a Kronecker sum:
\begin{equation}
  L = T_N \otimes I_N + I_N \otimes T_N,
  \label{eq:s_kronecker}
\end{equation}
where $I_N$ is the $N \times N$ identity and $\otimes$ denotes the Kronecker product. Because $T_N$ and $I_N$ commute in the Kronecker sense, both terms share the same eigenvectors, and $L$ is diagonalized by the 2D sine transform $S \otimes S$:
\begin{equation}
  L = (S \otimes S)\,(\Lambda \otimes I + I \otimes \Lambda)\,(S \otimes S)^{-1}.
  \label{eq:s_2d_diag}
\end{equation}
The eigenvalues of $L$ are therefore $\lambda_k + \lambda_l$ for $k,l = 1,\ldots,N$.

\paragraph{Spectral solve.}
Applying the 2D DST to both sides of Eq.~\eqref{eq:s_poisson} decouples the system into $N^2$ independent scalar equations:
\begin{equation}
  \hat{z}_{kl} = \frac{\hat{f}_{kl}}{\lambda_k + \lambda_l}, \qquad k,l = 1,\ldots,N,
  \label{eq:s_spectral_solve}
\end{equation}
where $\hat{z}$ and $\hat{f}$ are the 2D DST coefficients of $z$ and $f$. The depth map is recovered by the inverse 2D DST. Because the 2D DST is separable, it is computed as two successive 1D passes (row-wise, then column-wise), each of complexity $O(N \log N)$ per row (or column) when implemented via the FFT.

\paragraph{Note on DST type.}
The derivation above assumes a node-centred grid with boundary nodes $z_0 = z_{N+1} = 0$, yielding the DST-I basis functions $\sin(jk\pi/(N{+}1))$ and eigenvalues $\lambda_k = 2(\cos(k\pi/(N{+}1)) - 1)$ (Eq.~\eqref{eq:s_eigenval}). Our implementation instead uses the DST-II as defined in SciPy (\texttt{scipy.fft.dstn}, \texttt{type=2}), which corresponds to a half-pixel-shifted (cell-centred) grid where the boundary condition is imposed at half-grid-spacings beyond the first and last pixel. On this grid, the discrete Laplacian is diagonalized by the DST-II basis functions $\sin((2j{-}1)k\pi/(2N))$, with eigenvalues
\begin{equation}
  \lambda_k^{\mathrm{(II)}} = 2\!\left(\cos\frac{k\pi}{N} - 1\right), \qquad k = 1,\ldots,N.
  \label{eq:s_eigenval_dstII}
\end{equation}
For $k = 1$, $\lambda_1^{\mathrm{(II)}} \approx -\pi^2/N^2$ (matching the continuous eigenvalue in the limit); for $k = N$, $\lambda_N^{\mathrm{(II)}} = -4$. The spectral solve (Eq.~\eqref{eq:s_spectral_solve}) and 2D extension (Eqs.~\eqref{eq:s_kronecker}--\eqref{eq:s_2d_diag}) are algebraically identical under either convention; only the eigenvalue table stored in the hardware ROM differs. All precomputed reciprocals $1/(\lambda_k^{\mathrm{(II)}} + \lambda_l^{\mathrm{(II)}})$ are generated from Eq.~\eqref{eq:s_eigenval_dstII} and verified against the double-precision reference. For a comprehensive treatment of DST types and grid conventions, see Strang~\cite{strang2007cse}, Sections~7.2--7.3.

\end{document}